\documentclass{article}
\usepackage[preprint]{unireps_2025}

\usepackage[utf8]{inputenc} % allow utf-8 input
\usepackage[T1]{fontenc}    % use 8-bit T1 fonts
\usepackage{hyperref}       % hyperlinks
\usepackage{url}            % simple URL typesetting
\usepackage{booktabs}       % professional-quality tables
\usepackage{amsfonts}       % blackboard math symbols
\usepackage{nicefrac}       % compact symbols for 1/2, etc.
\usepackage{microtype}      % microtypography
\usepackage[dvipsnames]{xcolor} % colors
\usepackage{graphicx}
\usepackage{natbib}
\usepackage{makecell}
\usepackage{multirow}
\usepackage{amsmath}
\usepackage{siunitx, tabularx}
\usepackage{algorithm}
\usepackage{algpseudocode}
\usepackage{upgreek}

\title{GraphMatch: Fusing Language and Graph Representations in a Dynamic Two-Sided Work Marketplace}

\author{%
Mikołaj Sacha$^{1*}$ \quad Hammad Jafri$^1$ \quad Mattie Terzolo$^1$ \quad Ayan Sinha$^1$ \quad Andrew Rabinovich$^1$ \\
$^1$Upwork Inc.\\
$^*$\texttt{mikolajsacha@cloud.upwork.com}
}

\begin{document}

\maketitle

\begin{abstract}
 Recommending matches in a text-rich, dynamic two-sided marketplace presents unique challenges due to evolving content and interaction graphs. We introduce GraphMatch, a new large-scale recommendation framework that fuses pre-trained language models with graph neural networks to overcome these challenges. Unlike prior approaches centered on standalone models, GraphMatch is a comprehensive recipe built on powerful text encoders and GNNs working in tandem. It employs adversarial negative sampling alongside point-in-time subgraph training to learn representations that capture both the fine-grained semantics of evolving text and the time-sensitive structure of the graph. We evaluated extensively on interaction data from Upwork, a leading labor marketplace, at large scale, and discuss our approach towards low-latency inference suitable for real-time use. In our experiments, GraphMatch outperforms language-only and graph-only baselines on matching tasks while being efficient at runtime. These results demonstrate that unifying language and graph representations yields a highly effective solution to text-rich, dynamic two-sided recommendations, bridging the gap between powerful pretrained LMs and large-scale graphs in practice.
\end{abstract}

\section{Introduction}

Real-world recommendation systems, from e-commerce platforms and job or talent marketplaces to academic citation networks and news feeds, can be naturally described as \textit{Temporal Text-Attributed Graphs (TTAGs)}. In such TTAGs, nodes possess rich textual descriptions that evolve over time, while edges capture interactions like purchases, endorsements, or replies. Learning representations that effectively integrate \emph{both} the fine-grained semantics of text and the global, time-dependent structure of the graph is crucial for tasks such as matching, recommendation, and retrieval. This remains a significant open research challenge.

Recent advances in contrastive pretraining have led to language models (LM) such as Sentence-BERT~\cite{reimers2019sentencebert} and E5~\cite{wang2022text}. These models distill documents into single embeddings and excel at measuring \emph{local} semantic similarity.  However, two important signals are usually missing when LMs are applied to TTAGs. First, the decision to link two items often depends on the neighborhood they inhabit within the graph, an aspect that pure text encoders cannot see. Secondly, in TTAGs, the meaning of a node and the evidence for linking can change rapidly. For e.g., the rise of generative AI transformed "AI Writing" jobs posts suggesting editorial skills to prompt engineering ones. Furthermore, geopolitical events continuously influence the freelancer-client matching dynamics in the global work marketplace. Treating text snapshots as static features risks serious information leakage from future states.%{AS-> example in reference is non conventional.} (consider the dynamic "Apple" example in~\cite{zhang2025unifyingtextsemanticsgraph}). 

\paragraph{Why graph neural networks alone are not enough?}
Graph neural networks (GNNs) and their temporal variants are designed to capture long-range dependencies and temporal patterns. However, applying them directly to large-scale TTAGs is challenging. Each node might contain thousands of tokens, and the graph itself can have millions of time-stamped edges, with both text and structure evolving continuously. Jointly updating an LM and a GNN to handle this complexity often exceeds the memory and latency constraints of realistic applications~\cite{zhao2023learning, longa2023graphneuralnetworkstemporal}. Staged approaches, such as first fine-tuning a language model and then using its embeddings as static node features for a GNN~\cite{duan2023simteg}, can reduce computational overhead. However, they risk losing critical temporal signals, since static embeddings fail to capture evolving textual semantics and interaction patterns. This limitation is further amplified in \emph{two-sided marketplaces}, where matching quality depends on continuously balancing interests from distinct user groups.

\paragraph{Our contributions.}
We introduce \textbf{GraphMatch}, a framework designed to address these challenges. GraphMatch integrates three key design elements:

\begin{enumerate}
    \item \textbf{Scalable fusion of LMs and GNNs.} We begin by fine-tuning a Language Model (TextMatch) to derive domain-specific sentence embeddings. Then, GraphMatch employs these sentence embeddings using a lightweight residual projector. This strategy maintains a constant memory overhead per layer, enabling the GNN to iteratively refine representations by incorporating structural and temporal cues with rich textual semantics, thus enhancing scalability.
    \item \textbf{Adversarial negative sampling.} 
    We propose a novel approach to fine-tuning GNN with negative samples adversarial toward the LM embedding model. More specifically, after training TextMatch, we generate a training dataset for GraphMatch, which comprises recommendations generated by TextMatch with medium to high confidence. This sharpens the learning signal for GraphMatch, facilitating faster convergence toward improving the representations learned by TextMatch.
    \item \textbf{A practical recipe for TTAGs for two-sided user recommendation.} Drawing on effective strategies for large textual graphs, our training approach combines (i) temporally matched positive pairs from observed interactions, (ii) \emph{adversarially} mined negatives using TextMatch, and (iii) large in-batch negatives for efficient learning. We adapt our recipe for a two-sided marketplace recommendation system using task-homogeneous learning on mutually interesting objectives for both sides and effective handling of \emph{cold-start} nodes. This comprehensive approach promotes stable convergence on graphs with tens of millions of nodes.
\end{enumerate}

In this work, we present an extensive evaluation on a large-scale dataset derived user interaction data from Upwork~\cite{upwork}, a leading work marketplace platform, demonstrating that GraphMatch consistently outperforms strong LM-only and GNN-only baselines, as well as LM-and-GNN fusion without temporal aspect. %, while staying within operational latency budgets.
By offering a conceptually clear yet empirically powerful solution, GraphMatch advances representation learning on temporal, text-rich graphs in two sided marketplaces and provides a practical blueprint for future systems that need to reason about both language and evolving structure.

\begin{figure*}[t]
\centering
\includegraphics[width=\linewidth]{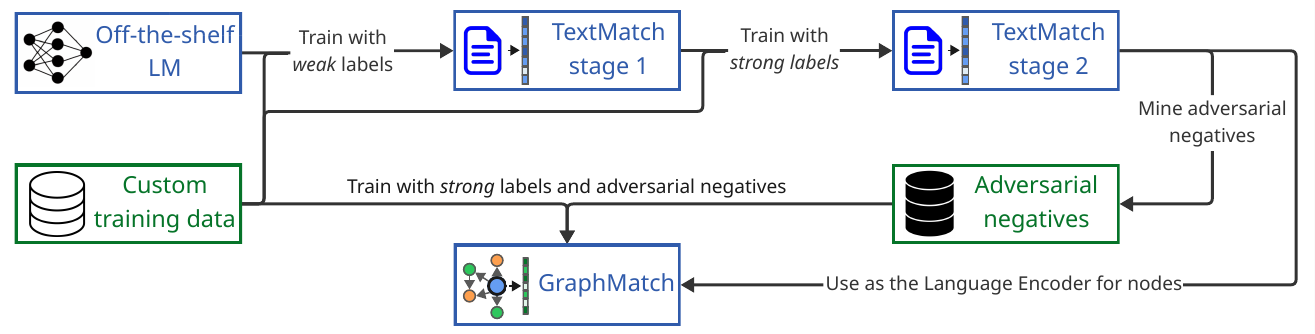}
\vspace{-0.4cm}
\caption{Flowchart of the multi-stage training of TextMatch and GraphMatch, with \textcolor{RoyalBlue}{models} and \textcolor{ForestGreen}{datasets} at different stages.}
\label{fig:training_flowchart}
\vspace{-0.3cm}
\end{figure*}

\section{Related Work}

Our research builds upon and extends several key areas within graph-based machine learning and recommendation systems. These include the application of Graph Neural Networks (GNNs) to recommendation tasks, techniques for learning representations from temporal and text-attributed graphs, the use of contrastive learning for graph representation, and the specific challenges of recommendation in two-sided marketplaces.

\paragraph{Graph Neural Networks for Recommendation.}
Graph Neural Networks (GNNs) have become a cornerstone in modern recommender systems due to their inherent ability to model complex relationships and capture collaborative signals within user-item interaction graphs~\cite{he2020lightgcn}. By iteratively passing and aggregating messages between nodes, GNNs can learn rich, structure-aware embeddings for users and items, effectively leveraging higher-order connectivity~\cite{hamilton2017inductive, velivckovic2017graph}. This structural modeling, combined with the ability to learn from node attributes, is beneficial for addressing challenges such as data sparsity and the cold-start problem within a recommendation ecosystem~\cite{Qian_2022}. Seminal works like GraphSAGE~\cite{hamilton2017inductive} and its variants have demonstrated the power of neighborhood sampling and aggregation for inductive representation learning on large graphs. Our work leverages these principles to learn representations within the context of a two-sided marketplace.

\paragraph{Learning on Temporal and Text-Attributed Graphs.}
Real-world graphs, especially in domains like online marketplaces, are often temporal, with evolving structures and node attributes, and text-attributed, where nodes carry rich semantic information. Learning effectively from such Temporal Text-Attributed Graphs (TTAGs) requires models that can jointly process structural, textual, and temporal dimensions~\cite{yan2023comprehensive, dtgb2024}. Common approaches to represent structure and text together involve using LMs to generate initial node features, which are then refined by a GNN that captures graph structure~\cite{duan2023simteg, chien2021node}. Other methods explore deeper fusion mechanisms between LMs and GNNs to better synergize semantic understanding with structural reasoning~\cite{yang2021graphformers, zhao2023learning, mavromatis2023train}. In contrast, GraphMatch proposes a scalable method to fuse LM-derived embeddings with a GNN that processes temporally-aware subgraphs while respecting textual semantics, an area of active research ~\cite{zhang2025unifyingtextsemanticsgraph, dtgb2024}.

\paragraph{Contrastive Learning for Graph Representations.}
Self-supervised learning, particularly through contrastive methods, has emerged as a powerful paradigm for learning graph representations without explicit labels~\cite{you2020graph, ju2023unsupervised, luo2023selfsupervised}. Graph Contrastive Learning (GCL) typically involves generating multiple views of a graph (or nodes) through augmentations~\cite{thakoor2022large}. The choice of augmentation strategies and the selection of negative samples are critical for the success of GCL \cite{trivedi2022augmentationsgraphcontrastivelearning, nozawa2021understanding, hafidi2022negative}. Techniques such as hard negative mining~\cite{robinson2020contrastive, kalanditis2020hard} and adaptive negative sampling~\cite{wang2023adans} aim to improve the quality of negative samples and thus the learned representations. GraphMatch utilizes contrastive objective in which empirically observed interactions are used as positive pairs, while TextMatch is used to mine adversarial hard negatives; this supply of temporally relevant positives and challenging negatives equips the GNN with precise signals that sharpen its representation learning.

% TODO: Cite the paper about no benchmarks for graph. highlight this is particularly true for two side marketplaces.
% TODO: Why time is so important in two sided marketplace? And it differentiates them. Network effect. Freelancers availablity is dynamic. Job posts are open for hires for a limited time.
\paragraph{Recommendation in Two-Sided Marketplaces.}
Two-sided marketplaces, such as freelance platforms~\cite{upwork} or e-commerce sites, present unique recommendation challenges that go beyond standard user-item matching~\cite{katukuri2013large, mehrotra2018towards}. These platforms must balance the preferences and constraints of two distinct groups of users, such as clients and freelancers. Effective matching in such marketplaces requires training methodologies that can capture the nuances of both sides~\cite{7033128}. GraphMatch tackles this challenge with a lightweight, three-part strategy: 
(i) \emph{temporal subgraph sampling} to track evolving interaction patterns 
(ii) \emph{task-homogeneous mini-batches} that isolate client-side and freelancer-side signals during training and 
(iii) a \emph{contrastive loss centered on “mutual-interest” events}, such as hires or interviews, which aligns the two sides, while we use TextMatch for adversarial hard negatives to improve discrimination.

\section{TextMatch}

We first use TextMatch to compute initial domain-adapted node representation denoted as $\Phi_\text{text\_emb}\!: (V \times \mathbb{R}) \rightarrow \mathbb{R}^{d_\text{TM}}$, which returns an embedding for any node $v \in V$ at any timestamp $t \in \mathbb{R}$, using only the textual content associated with the node. TextMatch embeddings serve as the initial semantic input for the heterogeneous graph processed by GraphMatch. Its core objective is to generate powerful vector representations for entities of any type.

\subsection{Model Architecture}
We employ a Sentence-BERT style~\cite{reimers2019sentencebert} encoder-only architecture. The parameters are shared between the query and candidate encoder, yielding a Symmetric Dual Encoder (SDE) model. This configuration is a common choice for top-performing text representation models on retrieval benchmarks such as BEIR~\cite{thakur2021beir} or MTEB~\cite{muennighoff2022mteb}.

\subsection{Two-Stage Domain Adaptation and Fine-Tuning}
% \subsection{Training Procedure}
To ensure that TextMatch embeddings are both generalizable and highly relevant to our specific domain, we adopt a two-stage training procedure (Figure~\ref{fig:training_flowchart}) based on established multi-stage recipes prevalent in state-of-the-art embedding models~\cite{ni2021largedualencodersgeneralizable, wang2022text}.

\textbf{Stage 1: Weakly Supervised Pre-fine-tuning.}
Initially, the model is adapted using naturally occurring textual pairs and \textit{weak} relevance signals sourced from the platform. Examples include \textit{profile-title $\leftrightarrow$ profile-body} associations and dense signals from user interactions within the marketplace, such as \textit{click} or \textit{save} events. Consistent with findings that such weak supervision, combined with large batch sizes and in-batch negatives, yields robust generic embeddings for retrieval~\cite{wang2022text}, we optimize temperature-scaled InfoNCE loss. In this stage, we prioritize large batch sizes and do not employ explicit hard-negative mining, allowing the model to learn broad semantic relationships while adapting to domain-specific vocabulary.

\textbf{Stage 2: Supervised Fine-Tuning.}
We refine the model's representations using stronger \emph{explicit} events as the primary supervisory signal. We use signals that show strong mutual interest between both sides of the marketplace, such as \textit{interviews} or \textit{hires}. For each positive pair, we incorporate a nuanced set of negaftives: (i) hard negatives, identified as chronologically adjacent impressions that \emph{have not} converted into a meaningful interaction, thus representing plausible yet rejected alternatives~\cite{robinson2020contrastive}, and (ii) random negatives sampled from the entire corpus to maintain distributional awareness. The InfoNCE loss is utilized again at this stage.

\paragraph{Training Stability and Two-Sided Marketplace Optimization.}
To optimally train for a two-sided marketplace, we construct \emph{task-homogeneous} mini-batches. Each batch contains examples exclusively from \emph{either} the client side \emph{or} the freelancer side of the marketplace. This strategy addresses challenges reported in contrastive learning where mixing heterogeneous tasks within a batch can degrade retrieval quality~\cite{hofstatter2021tasb}; specialized batching, such as the formation of single-task mini-batches, has been shown to mitigate this and improve the overall quality for model~\cite{SFRAIResearch2024}. % This batching strategy allows TextMatch to learn specialized representations for each side of the marketplace using events such as interviews as a strong, unifying signal of mutual interest.

\paragraph{Outcome.}
The TextMatch pipeline produces domain-adapted and semantically rich node embeddings. These embeddings serve as the initial features for GraphMatch. This allows GraphMatch to dedicate its capacity to modeling temporal and global graph structure instead of learning basic textual similarity, forming an essential step in the overall system. We present the quantitative improvements attributable to this approach in Section~\ref{sec:experiments}.

\section{GraphMatch}
\label{sec:graph_match}

\begin{figure}[t]
\includegraphics[width=\linewidth]{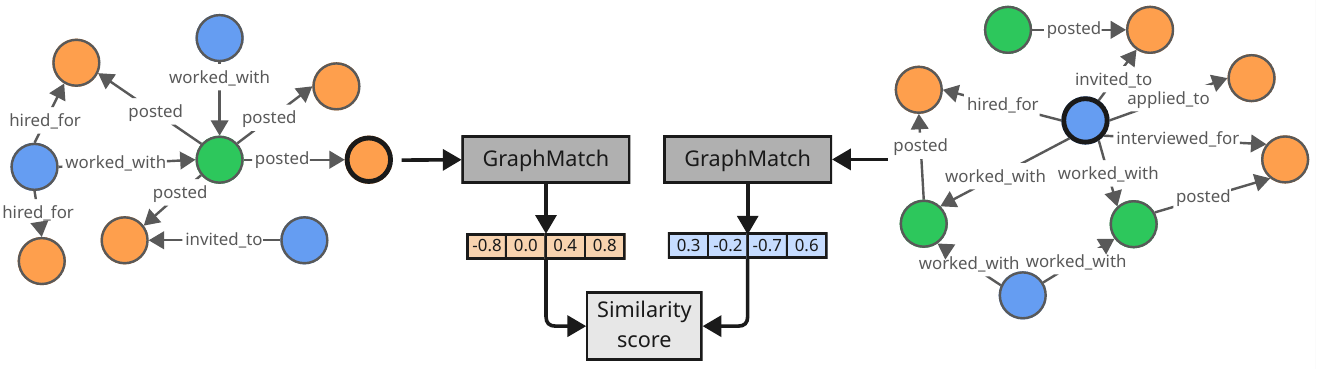}
\vspace{-0.6cm}
\caption{GraphMatch embeds \textcolor{RoyalBlue}{freelancers}, \textcolor{ForestGreen}{clients} or \textcolor{Orange}{job posts} using a sampled text-attributed graph representing their work history. In the illustration, we predict the embeddings of the emboldened job post (left) and freelancer (right) nodes using the surrounding graph. We compare GraphMatch embedding vectors using cosine similarity to predict the match probability between two entities.}
\label{fig:gnn_embedding_model}
\vspace{-0.1cm}
\end{figure}

\paragraph{Graph definition.} Similarly to TextMatch, GraphMatch learns numerical embeddings of entities in a user interaction graph. Let
\begin{equation}
G = (V, E, R, \mathcal{T}, \mathcal{A}, \Phi_\text{text\_emb}, \Phi_\text{feat}, \Psi)
\end{equation}
denote the user interaction graph where $V$ is the set of all nodes, $R$ is the set of possible interaction types, and $E \subseteq V \times V \times R$ is the set of all edges. We define $\mathcal{T}$ as the set of node types (for example, freelancers, job posts, clients). Each node $v \in V$ has a type $\tau(v) \in \mathcal{T}$. Each edge is associated with a timestamp with $\Psi: E \rightarrow \mathbb{R}$. 

We define $\mathcal{A}: V \rightarrow \mathcal{P}(\mathbb{R} \times \mathbb{R})$ as the activity periods function that maps each node $v \in V$ to a set of activity periods, where we represent each activity period as a tuple $(t_\text{start}, t_\text{end})$ with $t_\text{start} \leq t_\text{end}$. The set $\mathcal{A}(v)$ can be empty (indicating no recorded activity periods) or contain one or more non-overlapping time intervals during which node $v$ was \textit{active}. \textit{Activity period} may pertain to when a freelancer was open for work or when a job post was requesting applications.

\begin{figure*}[t]
\centering
\includegraphics[width=\linewidth]{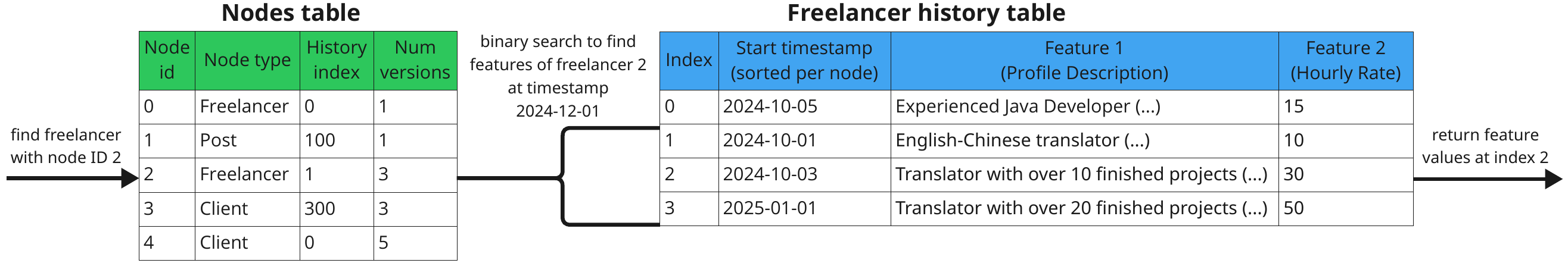}
%\caption{Node features in the work marketplace graph change over time. For example, a freelancer can change their profile description or hourly rate. We store node information in two tables. The main table (left) contains a single row per node, with its history start index and the number of versions. The node history table (right) stores all available versions of features per each node, grouped by node and sorted by timestamp within each node. Given any timestamp, node type and node ID, we first query the main table to retrieve the history index and the number of versions. Next, we run binary search over the relevant rows in the feature history table, which finds features in $O(log_{2}n)$ time. Finally, we return the point-in-time correct feature values. This efficient mechanism allows quickly retrieving features online from any arbitrary timestamp.}
\vspace{-0.5cm}
\caption{We store dynamic node features in two tables. The main table (left) contains a single row per node, with its history start index and the number of versions. The node history table (right) stores all available versions of features per each node, grouped by node and sorted by timestamp within each node. Given any timestamp, node type, and node ID, we first query the main table to retrieve the history index and the number of versions. Next, we run a binary search over the relevant rows in the feature history table, finding the point-in-time correct feature values in $O(log_{2}n)$ time.}
\label{fig:node_features_bin_search}
\vspace{-0.3cm}
\end{figure*}

$\Phi_\text{text\_emb}: (V \times \mathbb{R}) \rightarrow \mathbb{R}^{d_\text{TM}}$ maps every node and timestamp to its TextMatch embedding vector. We assume that TextMatch embedding is set to a zero vector for nodes without textual content. We store temporal TextMatch embedding versions per node in a sorted table as depicted in Figure~\ref{fig:node_features_bin_search}. For each node type $\uptau \in \mathcal{T}$, we define a type-specific feature function $\Phi_\text{feat}^\uptau: \{(v, t) \in V \times \mathbb{R} | \tau(v) = \uptau\} \rightarrow \mathbb{R}^{d_\uptau}$, where $d_\uptau$ is the dimension of numerical features for nodes of type $\uptau$. GraphMatch leverages this heterogeneous graph structure to learn an embedding function $f: V \rightarrow \mathbb{R}^{d_\text{GM}}$ that captures both textual similarity and interaction patterns to enhance matching performance.

\paragraph{Subgraph sampling.} In practice, using the full graph to compute node embeddings is computationally prohibitive. In Figure~\ref{fig:gnn_embedding_model}, we show how GraphMatch utilizes neighborhood subgraphs to compute entity embeddings. To construct the input graph for GraphMatch, given target node $v \in V$ and query timestamp $T \in \mathbb{R}$, we sample a temporal subgraph up to $K$ hops away from the target node. At each hop, for each edge type $r \in R$, we retain up to $N$ edges $e \in E$ such that $\Psi(e) \leq T$, selecting the $N$ most recent such edges if more than $N$ exist. Node features are also sampled as of timestamp $T$, using binary search over time-indexed feature snapshots to retrieve the most recent features prior to $T$ (see Figure~\ref{fig:node_features_bin_search}). This temporal subgraph construction enables GraphMatch to reason over temporally relevant interactions while maintaining efficiency.

\subsection{Model Architecture}
In this section, we describe the GraphMatch model. The GraphMatch framework is agnostic to some modeling choices such as the implementation of graph convolutional layer. We describe the key elements of our approach.
\paragraph{Initial Node Representation.}  
The initial representation of a node $v \in V$ depends on its type $\tau(v) \in \mathcal{N}$ and timestamp $t \in \mathbb{R}$. For each node type $n \in \mathcal{N}$, we have a type-specific encoder network $\text{Encoder}_n$ that processes the node features:

\begin{equation}
h_{v,t}^{(0)} = \text{Encoder}_{\tau(v)}
  \!\Bigl[\Phi_{\text{text\_emb}}(v, t)\,\|\,\Phi_{\text{feat}}^{\tau(v)}(v, t)\Bigr]
\end{equation}

Each encoder network $\text{Encoder}_n$ is a small feed-forward neural network that maps the node type-specific input features to a common embedding space.

\paragraph{Graph Convolution.}  
GraphMatch applies multiple layers of heterogeneous graph convolution to learn interaction-aware representations of nodes. Each heterogeneous graph convolution layer calculates
\[
h_{v,t}^{(l+1, r)} = \text{CONV}_r\left(h_{v,t}^{(l)}, \{h_{u,t}^{(l)} \mid u \in \mathcal{N}_r(v,t,K)\}\right),
\]
where \( h_{v,t}^{(l)} \) denotes the representation of node \( v \) at layer \( l \), \( \mathcal{N}_r(v,t,K) \) is the set of the K most recent neighbors of \( v \) connected by relation type \( r \) with edge timestamp $<t$, and \( \text{CONV}_r \) is a relation-specific aggregation function. The representations from each relation are aggregated (for example, by averaging) to obtain $h_{v,t}^{(l+1)}$. This process is repeated for multiple layers to capture higher-order interactions.
\paragraph{Final Embedding Fusion.}  
After the final graph convolution layer, the model applies a residual connection to the original TextMatch embedding:
\begin{equation}
f(v, t) = \frac{h_{v,t}^{\text{final}} + \text{Project}(\Phi_{\text{text\_emb}}(v,t))}{\left\|h_{v,t}^{\text{final}} + \text{Project}(\Phi_{\text{text\_emb}}(v,t))\right\|_2},
\end{equation}

where Project is a small feed-forward module that maps the TextMatch embedding to the final embedding space. This design ensures that the final embedding can benefit from TextMatch embeddings, numerical node features, and the graph-based learned representations.

\subsection{Training Procedure}
We train GraphMatch using a contrastive learning framework. The goal is to learn embeddings that place relevant query-candidate pairs closer in the embedding space while pushing away irrelevant pairs. 
\paragraph{Positive Training Labels.} We train GraphMatch for a temporal link prediction task. The set of training labels

\begin{equation}
C = \{(q_i, v_i^+, T_\text{start}, T_\text{end})\} \subseteq V \times V \times \mathbb{R} \times \mathbb{R}
\end{equation}

is derived from observed matches, where each match is associated with a relevance time span $(T_\text{start}, T_\text{end})$. For example, if a match represents a contract between a job post and a freelancer, the associated $T_\text{start}$ is the timestamp when the job offer was created, and $T_\text{end}$ is the timestamp when the contract was started. Note that this time span does not have to be equal to an activity time span from $\mathcal{A}$. For example, some job posts may request many freelancers and remain open for applications after a contract is initiated. For each sample during training, we randomly select a timestamp $t_i^+$ uniformly between $T_\text{start}$ and $T_\text{end}$ and make a prediction based on the version of the graph at $t_i^+$.

\paragraph{Adversarial Negative Mining.} 
We want GraphMatch to improve upon TextMatch embeddings by taking advantage of the user interaction graph. To achieve this, we employ negative samples generated \textit{adversarially} toward TextMatch. For each positive sample $(q_i, v_i^+, t_i^+)$, we generate up to $N$ adversarial negative samples by first building separate Approximate Nearest Neighbor search~\cite{annoy} indices for each node type using their respective activity periods, then performing similarity-based filtering to select challenging negatives. We present the algorithm pseudocode in Algorithm~\ref{alg:adv_neg}. In our experiments, we use $\sigma_\text{low}=0.5$, $\sigma_\text{high}=0.85$, $K=2000$, and $N=20$.

\begin{algorithm}[t]
\caption{Adversarial Negative Mining via TextMatch.}
\begin{algorithmic}[1]
\Require Positive dataset $(q_i, v_i^+, t_i^+) \in C \subseteq V \times V \times \mathbb{R}$
\Require TextMatch embedding function $\Phi_{\text{text\_emb}}: (V \times \mathbb{R}) \rightarrow \mathbb{R}^{d_{TM}}$
\Require Activity periods function $\mathcal{A}: V \rightarrow \mathcal{P}(\mathbb{R} \times \mathbb{R})$
\Require Node type function $\tau: V \rightarrow \mathcal{T}$
\Require Parameters: $K$, $N$, $\sigma_{\text{low}}$, $\sigma_{\text{high}}$

\State Build ANN indices $\mathcal{I}_{\mathcal{T}} \leftarrow \{\mathcal{I}_t : t \in \mathcal{T}\}$ where each $\mathcal{I}_t$ contains embeddings of nodes $v$ with $\tau(v) = t$ and active periods in $\mathcal{A}(v)$

\State $\mathcal{D}_{\text{train}} \leftarrow \emptyset$
\For{each $(q_i, v_i^+, t_i^+) \in C$}
    \State $e_{q_i} \leftarrow \Phi_{\text{text\_emb}}(q_i, t_i^+)$
    \State $\mathcal{C}_{K} \leftarrow \text{ANN}(e_{q_i}, K, 
    \mathcal{I}_{\tau(v_i^+)})$ \Comment{Nearest K within same node type}
    
    \State $\mathcal{F} \leftarrow \emptyset$
    \For{each $(v_j^-, t_\text{start}^-, t_\text{end}^-) \in \mathcal{C}_{K}$}
        \State $t_j^- \leftarrow \text{RandomUniform}(t_\text{start}^-, t_\text{end}^-)$
        \State $e_{v_j^-} \leftarrow \Phi_{\text{text\_emb}}(v_j^-, t_j^-)$
        \State $s_j^- \leftarrow \text{sim}(e_{q_i}, e_{v_j^-})$
        \If{$\sigma_{\text{low}} < s_j^- < \sigma_{\text{high}}$ and $v_j^- \neq v_i^+$}
            \State $\mathcal{F} \leftarrow \mathcal{F} \cup \{(v_j^-, t_j^-)\}$
        \EndIf
    \EndFor
    
    \State $\mathcal{N}_i \leftarrow \text{RandomSample}(\mathcal{F}, \min(N, |\mathcal{F}|))$
    \For{each $(v_j^-, t_j^-) \in \mathcal{N}_i$}
        \State $\mathcal{D}_{\text{train}} \leftarrow \mathcal{D}_{\text{train}} \cup \{(q_i, v_i^+, t_i^+, v_j^-, t_j^-)\}$
    \EndFor
\EndFor
\State \Return $\mathcal{D}_{\text{train}}$
\end{algorithmic}
\label{alg:adv_neg}
\end{algorithm}

Additionally, per each positive sample $(q_i, v_i^+, t_i^+)$, we generate a random negative candidate $(v_{i}^-, t_{i}^-)$ such that $\tau(v_{i}^-) = \tau(v_i^+)$ and $\exists (t_\text{start}, t_\text{end}) \in \mathcal{A}(v_{i}^-) : t_\text{start} \leq t_{i}^- \leq t_\text{end}$. This ensures that the space of negative samples covers all potential active nodes. In Section~\ref{sec:experiments}, we show that the combination of random and adversarial negatives is required to surpass the performance of TextMatch.

\paragraph{In-Batch Negative Sampling.} To enhance computational efficiency and exposure to diverse negative examples, we leverage in-batch negatives, similar to TextMatch. Assume our batch consists of queries $\{(q_i, t_i^+)\}$, positive candidates $\{(v_i^+, t_i^+)\}$ per each query and the set of all random and adversarial negative candidates generated for the batch $\{(v_j^-, t_j^-)\}$. We compute the similarities between each query $q_i$ and all positive and negative candidate entities in the batch:
\begin{align}
s^+_{i,j} &= f(q_i, t_i^+)^T f(v_j^+, t_j^+) \\
s^-_{i,j} &= f(q_i, t_i^+)^T f(v_j^-, t_j^-)
\end{align}
where $f(\cdot)$ denotes the GraphMatch embedding function.

\paragraph{Contrastive Loss.} We formulate the training objective as a cross-entropy loss where, for each query, the model must identify its corresponding positive candidate among all candidates in the batch:
\begin{equation}
\mathcal{L} = -\frac{1}{B}\sum_{i=1}^B \log\frac{\exp(s^+_{i,i}/\tau)}{\sum_{j \in C^+} \exp(s^+_{i,j}/\tau) + \sum_{j \in C^-} \exp(s^-_{i,j}/\tau)}
\end{equation}
where $B$ is the batch size, $i^+$ is the index of the positive candidate for query $i$, $C^+$ and $C^-$, respectively, represent the sets of all positive and negative candidate indices in the batch, and $\tau > 0$ is the temperature hyperparameter.

\section{Experiments}
\label{sec:experiments}
We train TextMatch and GraphMatch on two-sided work marketplace data. The training objective is to predict \textit{contracts} between freelancers and job posts based on freelancer profile overview and job description, and user interaction graph in the case of GraphMatch. We train TextMatch and GraphMatch in a multitask fashion, using training labels for predicting job posts for a query freelancer (FL $\rightarrow$ JP) or a freelancer for a query job post (JP $\rightarrow$ FL) with 50\%/50\% sampling probabilities. We depict the full training flowchart in Figure~\ref{fig:training_flowchart}.

\subsection{TextMatch Training}

The encoder is initialized from publicly available \texttt{E5-unsupervised} checkpoints (\textit{small}: $33$M and \textit{large}: $330$M parameters), which are themselves results of advanced contrastive pretraining~\cite{wang2022text}. We use the same encoder weights for queries and candidates. The pre-fine-tuning stage utilizes dense interaction data of ~110M samples based on \textit{weak} signals such as clicks or saves and naturally occurring pairs of titles and descriptions. The fine-tuning stage uses more sparse but \textit{strong} interactions, which show mutual interest between both sides of the marketplace, like interviews or a hires, combined with hard negatives based on the impressed but not interacted items mined from interaction logs.

\subsection{GraphMatch Training}

\paragraph{Training dataset.} The dataset for GraphMatch contains one year of user interaction data from Upwork~\cite{upwork}, with a 10/1/1 month temporal split for training, validation, and evaluation. The dataset consists of nodes representing freelancers, clients and job posts, and edges representing interactions such as posting a job by a client, job application, freelancer invitation, job interview, or contract start (see Figure~\ref{fig:gnn_embedding_model}). Freelancer and job post nodes are associated with textual features: profile overview and job description, respectively. All nodes contain additional numerical features specific to their type, such as freelancer hourly rate, client location, job post category, etc. Both text and numerical features can change over time. In total, there are approximately 9 million nodes with 32 million temporal versions of features and 62 million edges. We train GraphMatch on \textit{strong} labels derived from contracts between freelancers and job posts.

\paragraph{Training details.} We train GraphMatch using TextMatch-small embeddings for $\Phi_{text\_emb}$. The output embedding dimension for GraphMatch is $d_{GM} = 1024$. We freeze the weights of TextMatch and only update the weights of the GraphMatch-specific layers. We train GraphMatch for 24 hours, with hourly checkpointing, and for the final evaluation, select the checkpoint with the lowest validation loss. We use GATv2Conv~\cite{brody2021attentive} as the graph convolutional layer. In total, our model has approximately $64$M weights, including $33$M weights of TextMatch-small. We use a cluster of 8xNVIDIA A100 80GB GPUs for all training and evaluation.

\begin{table}[t]
% \small
\centering
\begin{tabular}{lrrr}
\toprule
\multirow{2}{*}{\textbf{Model}} & \multirow{2}{*}{\textbf{Adv Neg}} & \multicolumn{2}{c}{\textbf{NDCG@10}} \\
\cmidrule(lr){3-4}
& & \textbf{FL $\rightarrow$ JP} & \textbf{JP $\rightarrow$ FL} \\
\midrule
\midrule
snowflake-arctic-embed-l & & 7.4 \% & 6.1 \% \\
mxbai-embed-large-v1 &  & 11.2 \% & 6.7 \% \\
\midrule
TextMatch-small & & 21.6 \% & 10.6 \% \\
TextMatch-large & & 22.4 \% & 11.4 \% \\
\midrule
\multirow{2}{*}{GraphMatch-no-text} & $\times$ & 7.4 \% & 3.3 \% \\
& \checkmark & 12.8 \% & 7.4 \% \\
\midrule
\multirow{2}{*}{GraphMatch-no-feat} & $\times$ & 21.1 \% & 10.4 \% \\
& \checkmark & 22.4 \% & 11.8 \%  \\
\midrule
\multirow{2}{*}{GraphMatch-full} & $\times$ & 21.4 \% & 10.3 \%\\
& \checkmark & \textbf{24.2} \% & \textbf{12.4} \% \\
\bottomrule
\end{tabular}
\vspace{0.3cm}
\caption{Performance comparison on FL $\rightarrow$ JP and JP $\rightarrow$ FL retrieval, including GraphMatch trained with or without adversarial negatives.}
\label{tab:retrieval_metrics_comprehensive}
\vspace{-0.5cm}
\end{table}

\subsection{Evaluation}
\paragraph{Evaluation data.} The evaluation set comprises contracts between freelancers and job posts from 14 consecutive days within the evaluation part of the graph.  For each evaluation day, as the set of candidates, we gather all active freelancers and all job posts open for hire as of 12:00 GMT that day.  Out of these, we select as positive pairs the pairs of freelancers and job posts that engaged in a mutual contract with the contract start date after 12:00 GMT on that day. We evaluate for two mirrored retrieval tasks: retrieving the correct job post based on a query freelancer (FL $\rightarrow$ JP), and retrieving the correct freelancer based on a job post (JP $\rightarrow$ FL). For the JP $\rightarrow$ FL task, we only consider candidate freelancers who have completed their profile (profile completion > 80\%) and were active on the website within 48 hours prior to the considered timestamp. The dataset contains $\sim$15000 positive labels, with $\sim$13000 job posts per day and $\sim$24000 freelancers per day as the candidate pool.

\paragraph{Retrieval metrics.} In Table~\ref{tab:retrieval_metrics_comprehensive}, we show the aggregated NDCG@10 metric values for retrieval tasks FL $\rightarrow$ JP and JP $\rightarrow$ FL on the evaluation set. We compare TextMatch and GraphMatch with two state-of-the-art open-source embeddings models: \textit{snowflake arctic-embed-l}~\citep{merrick2024embeddingclusteringdataimprove} and \textit{mxbai-embed-large-v1}~\citep{embed2024mxbai}. We use the open-source model weights without any fine-tuning in a zero-shot fashion. We see that fine-tuning on the work marketplace dataset improves the accuracy of the text embedding model, exemplified by the high NDCG@10 of TextMatch compared to the open-source embedding models. The performance gap is significant despite the large size and generalized pre-training of the open source baselines, highlighting the importance of fine-tuning on the target dataset.

For GraphMatch, we observe that the combination of random and hard negatives is crucial for good performance. GraphMatch versions trained only with random negatives achieve performance comparable to TextMatch-small. We ablate over node features used by GraphMatch: \textit{no-feat} uses only TextMatch node embeddings, \textit{no-text} uses numerical node features without text embeddings, and \textit{full} uses the concatenation of TextMatch node embeddings and numerical node features. GraphMatch-full, trained with adversarial negatives, achieves the best performance, while GraphMatch-no-feat performs comparably to TextMatch-large, despite having a much smaller model size. We observe that textual context is crucial for accurate predictions, showcased by the low accuracy of GraphMatch-no-text.

\begin{table}[t]
\small
\centering
\begin{minipage}[t]{0.42\textwidth}
\centering
\begin{tabular}{p{2.2cm}rr}
\toprule
\multirow{2}{*}{\textbf{Model}} & \multicolumn{2}{c}{\textbf{NDCG@10}} \\
\cmidrule(lr){2-3}
& \textbf{FL $\rightarrow$ JP} & \textbf{JP $\rightarrow$ FL} \\
\midrule
\midrule
GraphMatch-full & \textbf{24.2} \% & \textbf{12.4} \% \\
\midrule
\makecell[l]{GraphMatch-full \\ \footnotesize (no temporal nodes)} & 15.1 \% & 8.5 \%  \\
\makecell[l]{GraphMatch-full \\ \footnotesize (no temporal graph)} & 13.3 \% & 6.9 \% \\
\bottomrule
\end{tabular}
\caption{Ablation study on disabling temporal sampling during training.}
\label{tab:retrieval_temporal}
\end{minipage}%
\hfill
\begin{minipage}[t]{0.5\textwidth}
\centering
\begin{tabular}{p{2.6cm}rrr} 
\toprule 
\multirow{2}{*}{\textbf{Model}} & \multicolumn{3}{c}{\textbf{NDCG@10 (FL $\rightarrow$ JP)}} \\ 
\cmidrule(lr){2-4}
& \makecell[r]{\textbf{Query} \\ \textbf{CS}} & \makecell[r]{\textbf{Cand} \\ \textbf{CS}} & \makecell[r]{\textbf{Both} \\ \textbf{CS}} \\ 
\midrule \midrule 
TextMatch-small & 21.6 \% & 23.4 \% & \textbf{23.3} \% \\ 
TextMatch-large & 21.7 \% & \textbf{24.1} \% & 22.7 \% \\ 
\midrule 
GraphMatch-no-text & 5.7 \% & 6.9 \% & 5.8 \% \\ 
GraphMatch-no-feat & 24.1 \% & 23.2 \% & 21.9 \% \\ 
GraphMatch-full & \textbf{25.3} \% & 23.0 \% & 22.8 \% \\
\bottomrule 
\end{tabular}
\caption{Evaluation on samples where query, ground truth candidate or both of these entities are \textit{cold start} (CS) nodes.} 
\label{tab:retrieval_cold_start_nodes}
\end{minipage}
\vspace{-0.3cm}
\end{table}

\paragraph{Importance of temporally accurate sampling.} Sampling point-in-time correct subgraphs and node features is necessary to train a model that generalizes well to unseen nodes. To verify this, we train GraphMatch in two setups. The \textit{no-temporal-nodes} version does not use temporally relevant node features during training (Figure~\ref{fig:node_features_bin_search}) but always utilizes the newest feature version in the training set. The \textit{no-temporal-graph} version additionally does not sample temporally relevant subgraphs but has access to all the graph edges during training. In Table~\ref{tab:retrieval_temporal}, we show that such sampling leads to poor generalization to the evaluation set, with much lower NDCG@10 values.

\paragraph{Retrieval accuracy on cold start nodes. } The \textit{cold start problem} is a critical challenge for recommender systems. In our setting, a cold start node can be a freelancer who has not yet interacted with other users or job posts, or a job post posted by a client who has not had any other interactions so far.%We illustrate the cold start nodes evaluation cases for the FL $\rightarrow$ JP retrieval task in Figure~\ref{fig:cold_start_nodes}.
We evaluate the performance of TextMatch and GraphMatch (trained with adversarial negatives) on the evaluation samples where the query or ground truth candidate entity is a cold start node. Table~\ref{tab:retrieval_cold_start_nodes} presents the evaluation results on samples for which query or ground truth candidate is a \textit{cold start} node. We see that the performance of GraphMatch on cold start nodes is comparable to TextMatch, indicating the model's ability to default to using text embeddings where no graph information is available, mitigating the cold start problem.

% Mikołaj: We can remove this figure if it takes up too much space. It is not super valuable.
%\begin{figure}[t]
%\includegraphics[width=\linewidth]{gnn_cold_start_examples.pdf}
%\caption{Examples of cold start target nodes for the FL $\rightarrow$ JP task. }
% A cold start freelancer has no interactions on the platform yet. A cold start job post is only attached to the posting client, who has no other interactions. 
%\label{fig:cold_start_nodes}
%\end{figure}

\section{Real-time Deployment of GraphMatch}
The real-time deployment architecture of GraphMatch consists of three main components: \textit{feature store}, \textit{inference service}, and \textit{graph database}. %(Figure~\ref{fig:service_flow_chart}).
This separation supports a shared, universal graph structure across various use cases while allowing individual GNN models to leverage customized node features. We implement node sampling using Cypher~\cite{cypher2018}, letting callers define the node and edge types needed at inference.

\paragraph{Feature Store.} Node features are primarily aggregated statistics (counts, averages, etc.). Our data warehouse, Snowflake~\cite{snowflake}, stores clones of production relational databases and assorted business data. Feature tables are generated via hourly SQL Mesh~\cite{sqlesh} ETL jobs and maintained in Feast~\cite{feast}. Since features update hourly, we define smart default values to handle missing features for entities present in the graph but not yet in the feature store.

\paragraph{Graph Database.} We host graph using Neo4j Aura~\cite{neo4jaura} graph database. Two ETL pipelines update the graph: a near real-time Kafka~\cite{kafka} pipeline for highly dynamic entities like job posts and an hourly Airflow~\cite{airflow} batch pipeline for other entities. In the graph database, we store minimal node properties (type, ID, creation timestamp) to enable accurate subgraph sampling for inference.
% When upstream services request embeddings for an entity that has not been loaded to the graph database, we simply return a single node since it is likely that node is not connected to any other nodes yet.

\paragraph{Inference Service.} Model inference runs on a Python/FastAPI~\cite{fastapi} microservice. We deploy the model on EC2 instances with Nvidia A10G GPUs (24GB RAM). Requests per second drive autoscaling, as GPU/CPU utilization is less predictive of latency degradation as usage increases. To optimize model inference, we compute and cache TextMatch embeddings for nodes every time a freelancer profile or job post description is updated, effectively limiting online inference to only the GNN layers. This enables achieving average embedding generation latencies of less than 70 ms per embedding across all entity types, supporting horizontal scalability and unlocking near real-time use cases.

%\begin{figure}[h]
%\centering
%\includegraphics[width=0.8\linewidth]{service_flowchart.pdf}
%\vspace{-0.6cm}
%\caption{Information flow of GraphMatch live inference.}
%\label{fig:service_flow_chart}
%\vspace{-0.0cm}
%\end{figure}

% TODO: we are handicapped because of no good benchmarks
% TODO: try to somehow highlight that it is generalizable even though we did not show it
\section{Conclusion}
In this work, we introduced \textbf{GraphMatch}, a framework for learning entity representations in a two-sided marketplace by integrating textual attributes and evolving user-interaction patterns using Temporal Text Attributed Graphs (TTAGs). Our experiments on a large-scale dataset from a real-world labor marketplace demonstrate that GraphMatch significantly outperforms text-only embedding approaches for matching freelancers with job posts. We show that effectively fusing textual and graph embeddings, with appropriate temporal sampling, achieves optimal performance. Additionally, we provided practical guidelines for deploying GraphMatch as a real-time, graph-based recommendation system.

Although GraphMatch is effective in recommendation tasks, it has few limitations. Its deployment complexity exceeds that of text-only embeddings, primarily because it requires real-time access to TTAGs during inference. Furthermore, the multi-stage training process of TextMatch and GraphMatch, while computationally expensive, is essential for accurately capturing textual, temporal, and structural features inherent to TTAGs.

Several directions for future research could enhance our approach. Firstly, incorporating a more sophisticated fusion mechanism, as suggested in~\cite{zhao2023learning,yang2021graphformers}, may further improve the integration of textual and graph embeddings. Secondly, exploring multi-stage training for GraphMatch akin to TextMatch could enhance embedding quality. Lastly, deeper investigations into node and edge type definitions or subgraph sampling strategies is interesting. Ultimately, although we focused on work marketplace, the principles and methodologies underpinning GraphMatch are broadly applicable to any two-sided marketplace. 

\clearpage 

\bibliography{unireps_2025}

%%%%%%%%%%%%%%%%%%%%%%%%%%%%%%%%%%%%%%%%%%%%%%%%%%%%%%%%%%%%

\clearpage 
\appendix

% \section*{Supplementary Materials}  % This creates an unnumbered section

%%%%%%%%%%%%%%%%%%%%%%%%%%%%%%%%%%%%%%%%%%%%%%%%%%%%%%%%%%%%

\end{document}